\documentclass[11pt]{article}

\usepackage[bookmarks,
            bookmarksopen = true,
            bookmarksnumbered = true,
            breaklinks = true,
            linktocpage,
            colorlinks = true,
            linkcolor = black,
            urlcolor  = black,
            citecolor = black,
            anchorcolor = black,
            hyperindex = true,
            hyperfigures
            ]{hyperref}
\usepackage{cite}
\usepackage{xcolor}
\usepackage{acl2015}
\usepackage{times}
\usepackage{url}
\usepackage{latexsym}
\usepackage{array}
\usepackage{multirow}
\usepackage{amssymb}
\usepackage{amsmath}
\usepackage{mathtools}
\usepackage{amsthm}
\usepackage{xspace}
\usepackage{algorithm}
\usepackage{tikz-dependency}
\usepackage{cancel}
\usepackage{balance}

\newcolumntype{P}[1]{>{\centering\arraybackslash}p{#1}}

\newcommand{\notes}[1]{}

\theoremstyle{definition}

\theoremstyle{plain}

\newcommand{\ith}[1]{\ensuremath{i^{{th}}}}

\newcommand{\ngram}{\ensuremath{\text{$n$-gram}}\xspace}
\newcommand{\ngrams}{\ensuremath{\text{$n$-grams}}\xspace}

\newcount\permx
\newcount\permy
\def\permdot#1#2{
\permx=#1 \advance\permx by-1
\permy=#2 \advance\permy by-1
\psframe[fillcolor=black, fillstyle=solid]
(\permx,\permy)(#1, #2)
}

\newcommand{\boxnum}[1]{{\setlength{\fboxsep}{1pt}\raisebox{1pt}{\hspace{1pt}\fbox{\tiny #1}\hspace{1pt}}}}
\newcommand{\ind}[1]{\ensuremath{_{\kern-0.5pt\boxnum{#1}}}}

\newcommand{\vecx}{\ensuremath{\mathbf{x}}\xspace}
\newcommand{\vecw}{\ensuremath{\mathbf{w}}\xspace}
\newcommand{\vecc}{\ensuremath{\mathbf{c}}\xspace}
\newcommand{\vecR}{\ensuremath{\mathbb{R}}\xspace}

\def\namecite{\newcite}

\newcommand{\smallnt}[1]{\ensuremath{_{\mbox{\tiny PP}}}\xspace}

\newcommand{\pseudocode}{Algorithm}
\floatname{algorithm}{\pseudocode}

\iffalse

\else

\fi

\newcommand{\dcnn}{DCNN\xspace}
\newcommand{\dcnns}{DCNNs\xspace}

\newcolumntype{P}[1]{>{\centering\arraybackslash}p{#1}}

\title{Dependency-based Convolutional Neural Networks\\for Sentence Embedding\thanks{
\; \scriptsize This work was done at both IBM and CUNY,
and was supported in part by DARPA FA8750-13-2-0041 (DEFT),
and NSF IIS-1449278.
We thank Yoon Kim for sharing his code,
and James Cross and Kai Zhao for discussions.} 
}

\iftrue
\author{
\vspace{0.1cm}
\hspace{0.2cm}Mingbo Ma$^{\dagger}$ \hspace{1cm} Liang Huang$^\dagger$ $^\ddagger$  \\
$^\dagger$Graduate Center \& Queens College\\ 
City University~of New York\\
{\tt \{mma2,lhuang\}gc.cuny.edu}
\And
\vspace{0.1cm}
\hspace{1.5cm}Bing Xiang$^\ddagger$   \hspace{1cm} Bowen Zhou$^\ddagger$\\
\hspace{1.6cm}$^\ddagger$IBM Watson Group\\
\hspace{1.6cm} T.~J.~Watson Research Center, IBM\\
\qquad {\tt \{lhuang,bingxia,zhou\}@us.ibm.com}
}
\fi

\date{}

\begin{document}
\maketitle
\begin{abstract}
In sentence modeling and classification, 
convolutional neural network approaches 
have recently achieved state-of-the-art results,
but all such efforts process word vectors sequentially and neglect long-distance dependencies.
To combine deep learning with linguistic structures,
we propose a dependency-based convolution approach, 
making use of tree-based $n$-grams rather than surface ones, 
thus utlizing non-local interactions between words. 
Our model improves sequential baselines 
on all four sentiment and question classification tasks, and 
achieves the highest published accuracy on TREC.
\end{abstract}

\vspace{-0.1cm}
\section{Introduction}
\vspace{-0.1cm}
Convolutional neural networks (CNNs),
originally invented in computer vision \cite{LeCun95comparisonof},
has recently attracted much attention in natural language processing (NLP) 
on problems such as sequence labeling \cite{collobert+:2011},
semantic parsing \cite{yih+:2104}, and search query retrieval \cite{yelong2014}.
In particular, recent work on CNN-based sentence modeling \cite{blunsom:2014,kim:2014}
has achieved excellent, often state-of-the-art, results on various classification tasks
such as sentiment, subjectivity, and question-type classification.
However, despite their celebrated success,
there remains a major limitation from the linguistics perspective:
CNNs, being invented on pixel matrices in image processing, 
only consider sequential \ngrams that are consecutive on the surface string 
and neglect long-distance dependencies,
while the latter play an important role
in many linguistic phenomena such as negation, subordination, and {\em wh}-extraction,
all of which might dully affect the sentiment, subjectivity, or 
other categorization of the sentence.

Indeed, in the sentiment analysis literature,
researchers have incorporated long-distance information from syntactic parse trees,
but the results are somewhat inconsistent:
some reported small improvements \cite{gamon:2004,Matsumoto:2005}, 
while some otherwise \cite{dave+:2003,Kudo2004}.
As a result, syntactic features have yet to become popular in the sentiment analysis community.
We suspect one of the reasons for this
is data sparsity (according to our experiments, tree \ngrams are significantly sparser than surface \ngrams), but this problem has largely been alleviated by the recent advances in word embedding.
Can we combine the advantages of both worlds? 

So we propose a very simple dependency-based convolutional neural networks (\dcnns). 
Our model is similar to \namecite{kim:2014},
but while his sequential CNNs put a word in its sequential context,
ours considers a word and its parent, grand-parent, great-grand-parent, and siblings on the dependency tree.
This way we incorporate long-distance information that are otherwise unavailable on the surface string.

Experiments on three classification tasks 
demonstrate the superior performance of our \dcnns
over the baseline sequential CNNs.
In particular, our accuracy on the TREC dataset
outperforms all previously published results in the literature,
including those with heavy hand-engineered features. 

Independently of this work, \namecite[unpublished]{mou+:2015} reported related efforts;
see Sec.~\ref{sec:related}. 

\begin{figure*}
\centering
\vspace{-0.4cm}
\includegraphics[width=0.8\textwidth]{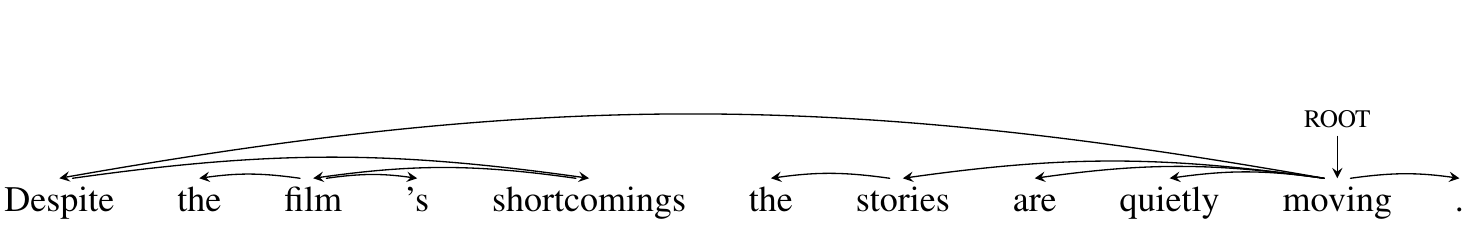}
\vspace{-0.4cm}
\caption{Dependency tree of an example sentence from the Movie Reviews dataset.}
\label{fig:example}
\vspace{-0.3cm}
\end{figure*}
\vspace{-0.1cm}
\section{Dependency-based Convolution}
\vspace{-0.1cm}
\noindent 
The original CNN, first proposed by \namecite{LeCun95comparisonof},
applies convolution kernels on a series of continuous areas of given images,
and was adapted to NLP by \namecite{collobert+:2011}.
Following \namecite{kim:2014}, 
one dimensional convolution operates the convolution kernel in sequential order
in Equation~\ref{eq:seq_con},
where~$\vecx_i \in \vecR^d$ represents the $d$ dimensional word representation 
for the $i$-th word in the sentence,
and~$\oplus$ is the concatenation operator. 
Therefore $\widetilde{\vecx}_{i,j}$ refers to concatenated word vector 
from the $i$-th word to the $(i+j)$-th word: 
\vspace{-2mm}
\begin{equation}\label{eq:seq_con}
    \widetilde{ \vecx}_{i,j} =   \vecx_i \oplus   \vecx_{i+1}\oplus \cdots \oplus  \vecx_{i+j} 
\vspace{-2mm}
\end{equation}

Sequential word concatenation $\widetilde{ \vecx}_{i,j}$ works as \ngram models 
which feeds local information into convolution operations. 
However, this setting can not capture long-distance relationships
unless we enlarge the window indefinitely which would inevitably cause the data sparsity problem.

In order to capture the long-distance dependencies
we propose the dependency-based convolution model (\dcnn).
Figure~\ref{fig:example} illustrates an example from the Movie Reviews (MR) dataset \cite{Pang+Lee:05a}. 
The sentiment of this sentence is obviously positive,
but this is quite difficult for sequential CNNs 
because many \ngram windows would include the highly negative word ``shortcomings'',
and the distance between ``Despite'' and ``shortcomings'' is quite long.
\dcnn, however, could capture the tree-based bigram ``Despite -- shortcomings'',
thus flipping the sentiment,
and the tree-based trigram ``ROOT -- moving -- stories'', which is highly positive.

\begin{figure*}
\centering
\includegraphics[width=0.8\textwidth]{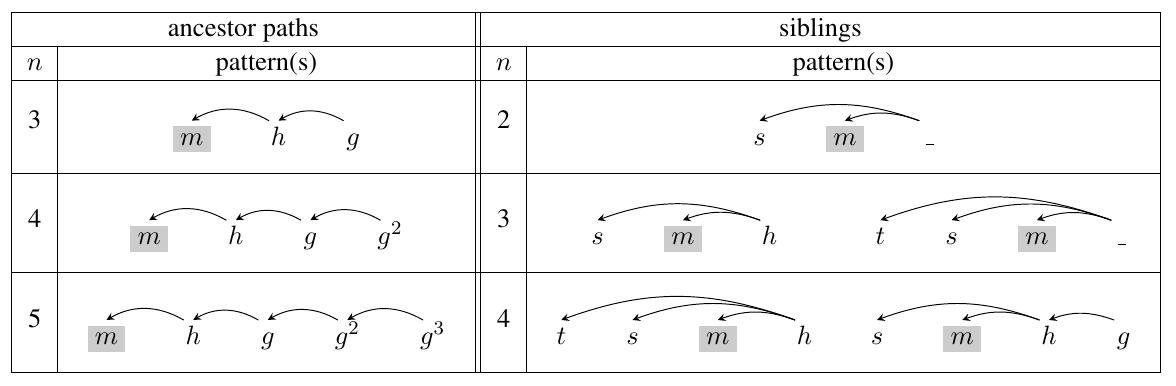}
\caption{Convolution patterns on trees. 
Word concatenation always starts with  $m$,
while $h$, $g$, and $g^2$ denote parent, grand parent, and great-grand parent, etc., 
and ``\_''  denotes words excluded in convolution.\label{tab:conv_features}}
\label{fig:conv_features}
\vspace{-0.5cm}
\end{figure*}

\vspace{-0.1cm}
\subsection{Convolution on Ancestor Paths}
\vspace{-0.1cm}
We define our concatenation based on the dependency tree for a given modifier $ \vecx_i$:
\vspace{-2mm}
\begin{equation}\label{eq:tree_con}
     \vecx_{i,k} =   \vecx_{i} \oplus   \vecx_{p(i)}\oplus \cdots \oplus  \vecx_{p^{k-1}(i)} 
\vspace{-2mm}
\end{equation}
where function $p^k(i)$ returns the $i$-th word's $k$-th ancestor index, which is recursively
defined as:
\vspace{-2mm}
\begin{equation}\label{eq:parent_def}
p^k(i)= \begin{cases}
 p(p^{k-1}(i)) & \text{if} \quad k>0 \\ 
 i             & \text{if} \quad k=0  \\
\end{cases}
\vspace{-2mm}
\end{equation}

Figure~\ref{fig:conv_features} 
(left) illustrates ancestor paths patterns with various orders. 
We always start the convolution with $x_i$ and concatenate with its ancestors. 
If the root node is reached, we add ``ROOT'' as dummy ancestors (vertical padding).

For a given tree-based concatenated word sequence $\vecx_{i,k}$, 
the convolution operation applies a filter $\mathbf{w} \in  \vecR^{k \times d}$ to $ \vecx_{i,k}$ with a bias term $b$ described in equation \ref{eq:con_def}:
\vspace{-2mm}
\begin{equation}\label{eq:con_def}
    c_i = f(\vecw \cdot  \vecx_{i,k}+b )
\vspace{-2mm}
\end{equation}
where $f$ is a non-linear activation function such as rectified linear unit (ReLu) or sigmoid function. 
The filter \vecw is applied to each word in the sentence,
generating 
the feature map $\vecc \in  \vecR^{l}$: 
\vspace{-2mm}
\begin{equation}\label{eq:featuremap}
    \vecc= [c_1, c_2, \cdots, c_l]
\vspace{-2mm}
\end{equation}
where $l$ is the length of the sentence.

\vspace{-0.1cm}
\subsection{Max-Over-Tree Pooling and Dropout}
\vspace{-0.1cm}
The filters convolve with different word concatenation in Eq.~\ref{eq:con_def} can be regarded as pattern detection: only the most similar pattern between the words and the filter 
could return the maximum activation. 
In sequential CNNs, max-over-time pooling \cite{collobert+:2011,kim:2014} operates over the feature map
to get the maximum activation $\hat{c} = \max {\vecc}$ representing the entire feature map. 
Our \dcnns also pool the maximum activation from feature map 
to detect the strongest activation over the whole tree (i.e., over the whole sentence).
Since the tree no longer defines a sequential ``time'' direction,
we refer to our pooling as ``max-over-tree'' pooling.

In order to capture enough variations, we randomly initialize the set of filters to detect different structure patterns. Each filter's height is the number of words considered and the width is always equal to the dimensionality $d$ of word representation. 
Each filter will be represented by only one feature after max-over-tree pooling. 
After a series of convolution with different filter with different heights, multiple features carry different structural information become the final representation of the input sentence. 
Then, this sentence representation is passed to a fully connected soft-max layer and outputs a distribution over different labels.

Neural networks often suffer from overtraining. 
Following \namecite{kim:2014},
we employ random dropout on penultimate layer \cite{hinton:2014}. 
in order to prevent co-adaptation of hidden units. 
In our experiments, we set our drop out rate as 0.5 and learning rate as 0.95 by default. 
Following \namecite{kim:2014}, training is done through stochastic gradient descent 
over shuffled mini-batches with the
Adadelta update rule \cite{zeiler:2012}.

\vspace{-0.1cm}

\subsection{Convolution on Siblings}
\vspace{-0.1cm}
Ancestor paths alone is not enough to capture many linguistic phenomena such as conjunction.
Inspired by higher-order dependency parsing \cite{mcdonald+pereira:2006,koo+collins:2010},
we also incorporate siblings for a given word in various ways. 
See Figure~\ref{fig:conv_features} 
(right) for details.

\vspace{-0.1cm}

\subsection{Combined Model}
\vspace{-0.1cm}
Powerful as it is, structural information still does not fully cover 
sequential information.
Also, parsing errors (which are common especially for informal text such as online reviews) 
directly affect \dcnn performance 
while sequential \ngrams are always correctly observed.
To best exploit both information, we want to combine both models.
The easiest way of combination is to concatenate these representations together, then feed into fully connected soft-max neural networks. 
In these cases, combine with different feature from different type of sources could stabilize the performance. 
The final sentence representation is thus:
\[
\hat{\vecc} = [\underbrace{\hat{c}_a^{(1)}, ..., \hat{c}_a^{(N_a)}}_\text{ancestors}; 
               \underbrace{\hat{c}_s^{(1)}, ..., \hat{c}_s^{(N_s)}}_\text{siblings};
               \underbrace{\hat{c}^{(1)}, ..., \hat{c}^{(N)}}_\text{sequential}]
\]
where $N_a$, $N_s$, and $N$ are the number 
of ancestor, sibling, and sequential filters.
In practice, we use 100 filters for each template in Figure~\ref{fig:conv_features} .
The fully combined representation is 1,100-dimensional
by contrast to 300-dimensional for sequential CNN.

\begin{table*}
\centering
\scalebox{0.9}{
\begin{tabular}{ |l|l||l|l|l|l|l| }
\hline
		Category 						&Model                             &   MR          &   SST-1     &   TREC & TREC-2\\ 
\hline
\multirow{3}{*}{This work } 
					       &\dcnns: ancestor                              & 80.4$^{\dagger}$          &  47.7$^{\dagger}$        &   95.4$^{\dagger}$        & 88.4$^{\dagger}$\\
                                  &\dcnns: ancestor+sibling           &  81.7$^{\dagger}$        &  48.3$^{\dagger}$       & \textbf{95.6}$^{\dagger}$  &  89.0$^{\dagger}$\\
                      &\dcnns: ancestor+sibling+sequential         &\textbf{81.9}&  49.5       &    95.4$^{\dagger}$ &  88.8$^{\dagger}$\\
\hline
\multirow{2}{*}{CNNs} 	  
                            & CNNs-non-static \cite{kim:2014} -- baseline           &  81.5          &  48.0         &    93.6  &    86.4$^*$  \\
 				      & CNNs-multichannel  \cite{kim:2014}         &  81.1         &  47.4         &    92.2    &   86.0$^*$   \\ 
					& Deep CNNs  \cite{blunsom:2014}               &  -              &  48.5         &    93.0     &  - \\

\hline
\multirow{2}{*}{Recursive NNs} 	  
  & Recursive Autoencoder \cite{SocherEtAl2011:RAE}                &  77.7        &  43.2          &    -     & -   \\
& Recursive Neural Tensor  \cite{Socher-etal:2013}     &  -               &  45.7          &    -     & -   \\ 
& Deep Recursive NNs  \cite{irsoy-drsv}                                     &  -             & \textbf{49.8}&    -    & -    \\ 
\hline
\multirow{1}{*}{Recurrent NNs} 	  
                                       & LSTM on tree \cite{xiaodan}             &\textbf{81.9}&  48.0          &    -    & -    \\

\hline
\multirow{1}{*}{Other deep learning} 	  
                            & Paragraph-Vec \cite{le2014distributed}         & -       & 48.7          &    -     & -   \\
\hline
\multirow{1}{*}{Hand-coded rules} 	
 				 & $\textup{SVM}_S$ \cite{silva11} &  -  &           &   95.0     & {\bf 90.8}  \\ 

\hline
\end{tabular}
}
\caption{Results on Movie Review (MR), Stanford Sentiment Treebank (SST-1), 
and TREC datasets. TREC-2 is TREC with fine grained labels.   
$^{\dagger}$Results generated by GPU (all others generated by CPU). 
$^*$Results generated from \namecite{kim:2014}'s implementation.}
 \label{tb:results}
\vspace{-0.5cm}
\end{table*}

\vspace{-0.2cm}
\section{Experiments}
\vspace{-0.15cm}
\noindent
Table~\ref{tb:results} summarizes  results
in the context of other high-performing efforts in the literature.
We use three benchmark datasets in two categories:
sentiment analysis on both Movie Review (MR) \cite{Pang+Lee:05a} 
and Stanford Sentiment Treebank (SST-1) \cite{Socher-etal:2013} datasets,
and question classification on TREC \cite{Li:2002:LQC:1072228.1072378}.

For all datasets, we first obtain the dependency parse tree from Stanford parser \cite{manning-EtAl:2014:P14-5}.\footnote{\scriptsize The phrase-structure trees in SST-1 are actually automatically parsed,
and thus can not be used as gold-standard trees.}
Different window size for different choice of convolution are shown in Figure~\ref{fig:conv_features}. 
For the dataset without a development set (MR), we randomly choose 10$\%$ of the training data 
to indicate early stopping. 
In order to have a fare comparison with baseline CNN, we also use 3 to 5 as our window size. 
Most of our results are generated by GPU due to its efficiency, however CPU could potentially get better results.\footnote{\scriptsize GPU only supports {\tt float32} while CPU supports {\tt float64}.} 
Our implementation, on top of 
\namecite{kim:2014}'s code,\footnote{\scriptsize \url{https://github.comw/yoonkim/CNN_sentence}}
will be released.\footnote{\scriptsize \url{https://github.com/cosmmb/DCNN}} 
\vspace{-0.1cm}
\subsection{Sentiment Analysis}
\vspace{-0.1cm}
\noindent
Both sentiment analysis datasets (MR and SST-1) are based on movie reviews. 
The differences between them are mainly in the different numbers of categories 
and whether the standard split is given. There are 10,662 sentences in the MR dataset. 
Each instance is labeled positive or negative,
and in most cases contains one sentence.
Since no standard data split is given, 
following the literature we use 10 fold cross validation to include every sentence in training and testing at least once. 
Concatenating with sibling and sequential information obviously improves DCNNs,
and the final model outperforms the baseline sequential CNNs by 0.4,
and ties with \namecite{xiaodan}. 

Different from MR, the Stanford Sentiment Treebank (SST-1) annotates finer-grained labels, very positive, positive, neutral, negative and very negative, on an extension of the MR dataset. 
There are 11,855 sentences with standard split. 
Our model achieves an accuracy of 49.5 which is second only to 
\namecite{irsoy-drsv}. 

\vspace{-0.1cm}
\subsection{Question Classification}
\vspace{-0.1cm}
In the TREC dataset, the entire dataset of 5,952 sentences 
are classified into the following 6 categories: abbreviation, entity, description, location and numeric.  
In this experiment, \dcnns easily outperform any other methods even with ancestor convolution only. 
\dcnns with sibling achieve the best performance in the published literature. \dcnns combined with sibling and sequential information might suffer from overfitting on the training data based on our observation. One thing to note here is that our best result even exceeds  $\text{SVM}_S$ \cite{silva11} with 60 hand-coded rules. 

\begin{figure}
\centering
\includegraphics[width=0.45\textwidth]{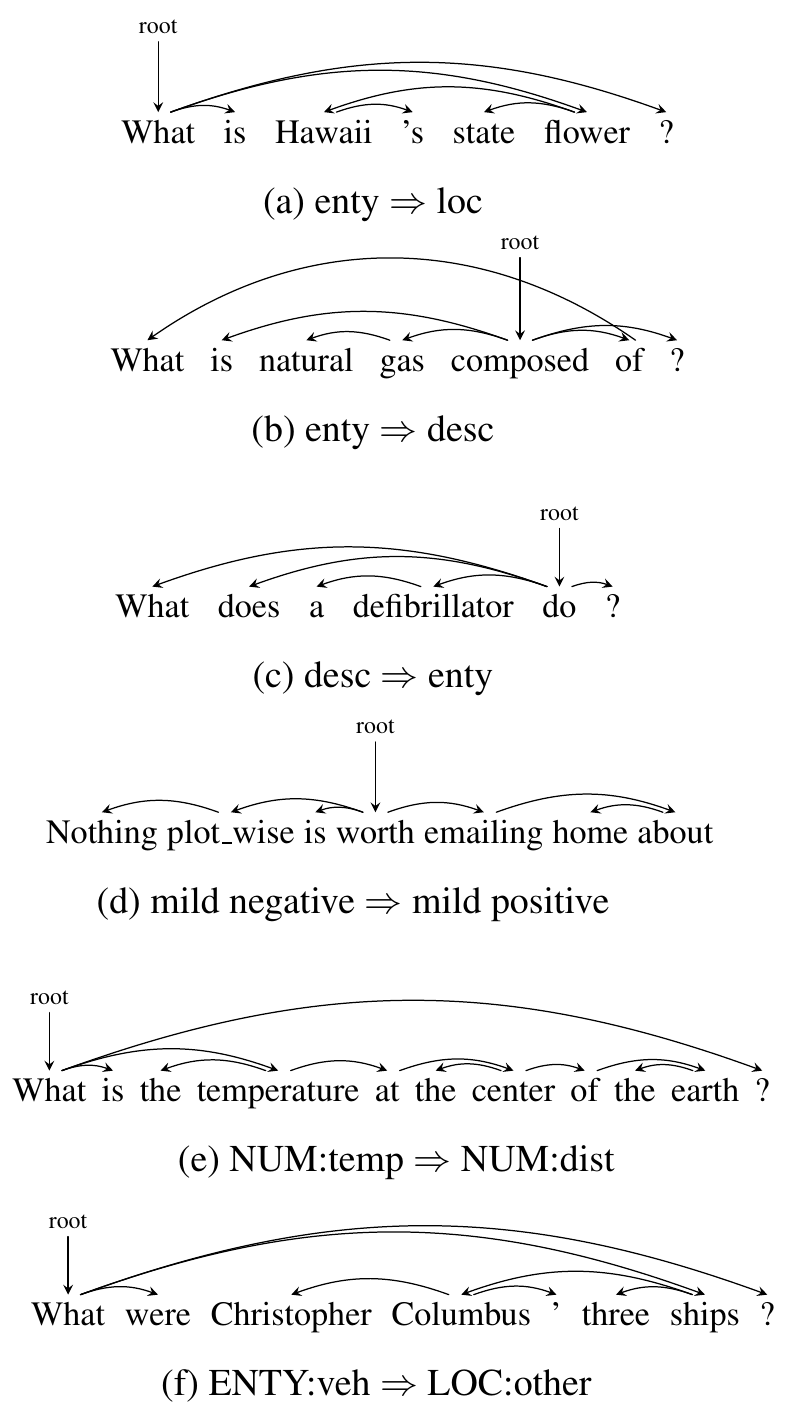}
\caption{Examples from TREC (a--c), SST-1 (d) and TREC with fine-grained label (e--f) 
that are misclassified by the baseline CNN but 
correctly labeled by our \dcnn.
For example, (a) should be {\em entity} but is 
labeled {\em location} by CNN.}
\label{fig:TRECandSST}
\vspace{-0.5cm}
\end{figure}

The TREC dataset also provides subcategories such as numeric:temperature, numeric:distance, and entity:vehicle. 
To make our task more realistic and challenging, 
we also test the proposed model with respect to the 50 subcategories.
There are obvious improvements over sequential CNNs from the last column of Table~\ref{tb:results}. 
Like ours, \namecite{silva11} is a tree-based system
but it uses constituency trees compared to ours dependency trees.
They report a higher fine-grained accuracy of 90.8 but 
their parser is trained {\em only} on the QuestionBank~\cite{questionbank:2006} 
while we used the standard Stanford parser trained on both the Penn Treebank 
and QuestionBank. 
Moreover, as mentioned above,
their approach is rule-based while ours is automatically learned.

\begin{figure}
\centering
\includegraphics[width=0.4\textwidth]{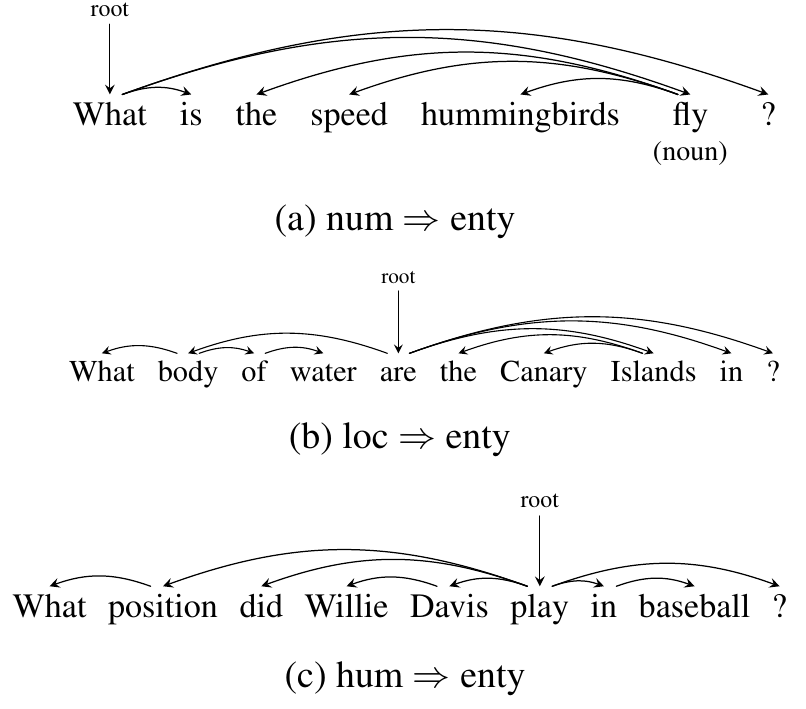}
\caption{Examples from TREC datasets
that are misclassified by \dcnn but 
correctly labeled by baseline CNN.
For example, (a) should be {\em numerical} but is 
labeled {\em entity} by \dcnn.}
\label{fig:ourmistakes}
\vspace{-0.5cm}
\end{figure}

\vspace{-0.1cm}
\subsection{Discussions and Examples}
\vspace{-0.1cm}

Compared with sentiment analysis, the advantage of our proposed model is 
obviously more substantial on the TREC dataset. 
Based on our error analysis, we conclude that this is mainly due to the difference of the parse tree quality between the two tasks. In sentiment analysis, the dataset is collected from the {\em Rotten Tomatoes} website which includes many irregular usage of language. 
Some of the sentences even come from languages other than English. 
The errors in parse trees inevitably affect the classification accuracy.
However, the parser works substantially better on the TREC dataset 
since all questions are in formal written English,
and the training set for Stanford parser\footnote{
\tiny\url{http://nlp.stanford.edu/software/parser-faq.shtml}}
already includes the QuestionBank \cite{questionbank:2006}
which includes 2,000 TREC sentences.


Figure~\ref{fig:TRECandSST} visualizes examples where
CNN errs while \dcnn does not.
For example, CNN labels (a) as {\em location} due to ``Hawaii'' and ``state'', 
while the long-distance backbone ``What -- flower'' is clearly asking for an {\em entity}.
Similarly, in~(d), \dcnn captures the obviously negative tree-based trigram ``Nothing -- worth -- emailing''.
Note that our model also works with non-projective dependency trees
such as the one in (b). The last two examples in Figure~\ref{fig:TRECandSST} visualize cases 
where \dcnn outperforms the baseline CNNs in fine-grained TREC. In example (e), the word ``temperature'' is at second from the top and is root of a 8 word span ``the ... earth''. When we use a window of size 5 for tree convolution, every words in that span get convolved with ``temperature'' and 
this should be the reason why \dcnn get correct. 

Figure~\ref{fig:ourmistakes} showcases examples where baseline CNNs get better results than \dcnns. Example (a) is misclassified as {\em entity} by \dcnn due to parsing/tagging error
(the Stanford parser performs its own part-of-speech tagging). 
The word ``fly'' at the end of the sentence should be a verb instead of noun,
and ``hummingbirds fly'' should be a relative clause modifying ``speed''.

\begin{figure}
\centering
\includegraphics[width=0.4\textwidth]{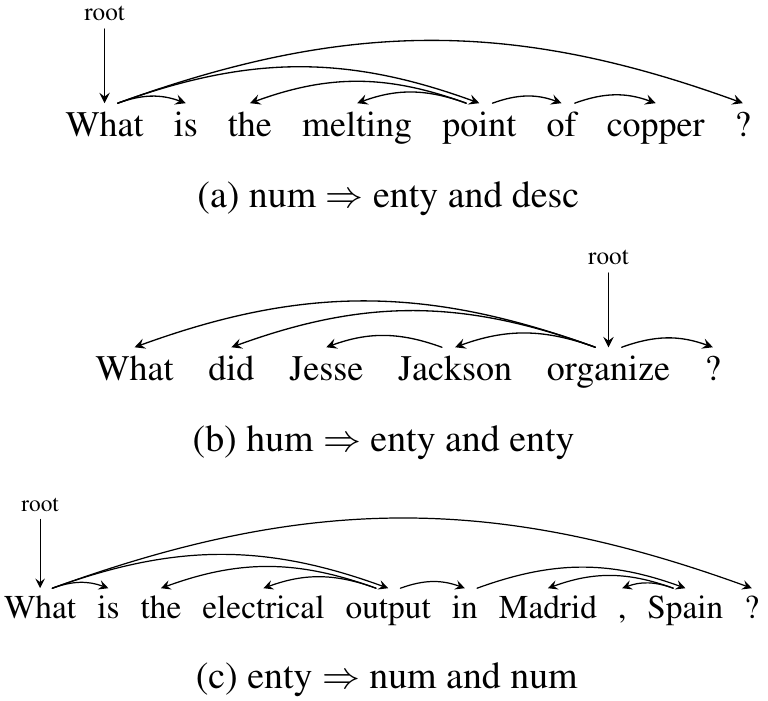}
\caption{Examples from TREC  datasets
that are misclassified by both \dcnn and baseline CNN.
For example, (a) should be {\em numerical} but is 
labeled {\em entity} by \dcnn and {\em description} by CNN.}
\label{fig:bothmistakes}
\vspace{-0.5cm}
\end{figure}

There are some sentences that are misclassified by both the baseline CNN and \dcnn. 
Figure~\ref{fig:bothmistakes} shows three such examples. Example (a) is not classified as {\em numerical} by both methods due to the ambiguous meaning of the word ``point'' which is difficult to capture by word embedding. This word can mean location, opinion, etc. Apparently, the numerical aspect is not captured by word embedding. Example (c) might be an annotation error. 


 \label{sec:related}

Shortly before submitting to ACL 2015 we learned 
\namecite[unpublished]{mou+:2015} 
have independently reported concurrent and related efforts.
Their constituency model,
based on their unpublished work in programming languages \cite{mou+:2014},\footnote{\scriptsize 
Both their 2014 and 2015 reports proposed
(independently of each other and independently of our work)
 the term ``tree-based convolution'' (TBCNN).
}
performs convolution on pretrained recursive {\em node} representations
rather than {\em word} embeddings, 
thus baring little, if any, resemblance to our dependency-based model.
Their dependency model is related, but always includes a node and all its children
(resembling \namecite{iyyer+:2014}), 
which is a variant of our sibling model and always flat.
By contrast, our ancestor model looks at the vertical path from 
any word to its ancestors, being linguistically motivated \cite{shen+:2008}.


\vspace{-0.2cm}
\section{Conclusions} 
\vspace{-0.2cm}
We have presented a very simple dependency-based convolution framework 
which outperforms sequential CNN baselines on modeling sentences.

\bibliographystyle{acl}
\balance
\bibliography{thesis}

\end{document}